\pdfoutput=1
\documentclass[11pt]{article}

\usepackage{EMNLP2022}
\usepackage{times}
\usepackage{latexsym}

\usepackage[T1]{fontenc}
\usepackage[utf8]{inputenc}

\usepackage{microtype}

\usepackage{inconsolata}

\usepackage{microtype}
\usepackage{amsmath}
\usepackage{amssymb}
\usepackage{multirow}
\usepackage{booktabs}

\usepackage{graphicx}
\usepackage{epstopdf}

\usepackage{amsmath}
\usepackage{CJKutf8}
\usepackage{url}

\title{Findings of the WMT 2022 Shared Task on Translation Suggestion}

\author{Zhen Yang,  Fandong Meng, Yingxue Zhang, Ernan Li, and Jie Zhou \\
   Pattern Recognition Center, WeChat AI, Tencent Inc, Beijing, China \\
  {\tt \{zieenyang, fandongmeng, yxuezhang, cardli, withtomzhou\}@tencent.com}}
  
\begin{document}
\begin{CJK}{UTF8}{gbsn} 
\maketitle
\begin{abstract}
We report the result of the first edition of the WMT shared task on Translation Suggestion (TS). The task aims to provide alternatives for specific words or phrases given the entire documents generated by machine translation (MT). It consists two sub-tasks, namely, the naive translation suggestion and translation suggestion with hints. The main difference is that some hints are provided in sub-task two, therefore, it is easier for the model to generate more accurate suggestions. For sub-task one, we provide the corpus for the language pairs English-German and English-Chinese. And only English-Chinese corpus is provided for the sub-task two.

We received 92 submissions from 5 participating teams in sub-task one and 6 submissions for the sub-task 2, most of them covering all of the translation directions. We used the automatic metric BLEU for evaluating the performance of each submission.

\end{abstract}

\section{Introduction}
Computer-aided translation (CAT) \cite{barrachina2009statistical,green2014human,knowles2016neural,santy2019inmt} has attained more and more attention for its promising ability in combining the high efficiency of machine translation (MT) \cite{cho2014learning,bahdanau:14,vaswani2017attention} and high accuracy of human translation (HT). A typical way for CAT tools to combine MT and HT is PE \cite{green2013efficacy,zouhar2021neural}, where the human translators are asked to provide alternatives for the incorrect word spans in the results generated by MT. To further reduce the post-editing time, researchers propose to apply TS into PE, where TS provides the sub-segment suggestions for the annotated incorrect word spans in the results of MT, and their extensive experiments show that TS can substantially reduce translators' cognitive loads and the post-editing time \cite{wang2020touch,lee2021intellicat}.

As there is no explicit and formal definition for TS, we observe that some previous works similar or related to TS have been proposed \cite{alabau2014casmacat,santy2019inmt,wang2020touch,lee2021intellicat}. However, there are two main pitfalls for these works in this line. First, most conventional works only focus on the overall performance of PE but ignore the exact performance of TS. This is mainly because the golden corpus for TS is relatively hard to collect. As TS is an important sub-module in PE, paying more attention to the exact performance of TS can boost the performance and interpretability of PE. Second, almost all of the previous works conduct experiments on their in-house datasets or the noisy datasets built automatically, which makes their experiments hard to be followed and compared. Additionally, experimental results on the noisy datasets may not truly reflect the model's ability on generating the right predictions, making the research deviate from the correct direction. Therefore, the community is in dire need of a benchmark for TS to enhance the research in this area. To address the limitations mentioned above and spur the research in TS, we make our efforts to construct a high-quality benchmark dataset with human annotation, named \emph{WeTS},\footnote{\emph{WeTS}: We Establish a benchmark for Translation Suggestion} which covers four different translation directions. 

The main motivation of this shared task is two-fold. The first goal is to analyze the challenges in the area of TS, which can provide some new directions for the further researches and applications in this area. Secondly, we want to make the researchers notice the gaps between the golden and automatically generated synthetic corpus. And we want to see the performance of different techniques on the golden corpus. As the source and translation sentence are both the inputs of TS, it is interesting to see how the interactions between the source and translation sentences can improve the final suggestions.

In order to evaluate the quality of the participating systems, we use the automatic metric, BLEU \cite{papineni2002bleu}. Specifically, we adopt the widely used toolkit, sacrebleu \cite{post-2018-call} to calculate the BLEU score for the top-1 suggestion against the reference sentences.\footnote{https://github.com/mjpost/sacrebleu} For Chinese, the BLEU score is calculated on teh character with the default tokenizer for Chinese. As for English, the BLEU score is calcualted on the case-sensitive words with the default tokenizer 13a.

Five teams participated in this first campaign of the Translation Suggestion shred task, most of them cover the four translation directions. We will describe each system which submits the technical paper in detail.

\section{Task Description}
This section describes the task definition in the first edition of TS shared task.
We finely divide the task of TS into two sub-tasks, namely \emph{vanilla TS} and \emph{TS with hints}, according to whether the translators' hints are considered.
\paragraph{Vanilla TS.} Given the source sentence $\boldsymbol{x}=(x_1, \ldots, x_s)$, the translation sentence $\boldsymbol{m}=(m_1,\ldots,m_t)$, the incorrect words or phrases $\boldsymbol{w}=\boldsymbol{m}_{i:j}$ where $1 \le i \le j \le t$, and the correct alternative $\boldsymbol{y}$ for $\boldsymbol{w}$, the task of \emph{vanilla TS} is optimized to maximize the conditional probability of $\boldsymbol{y}$ as follows:
\begin{equation}
    \label{eq: definition}
    P(\boldsymbol{y}|\boldsymbol{x}, \boldsymbol{m}^{-\boldsymbol{w}}, \theta)
\end{equation}
where $\theta$ represents the model parameter, and $\boldsymbol{m}^{-\boldsymbol{w}}$ is the masked translation where the incorrect word span $\boldsymbol{w}$ is replaced with a placeholder. \footnote{$\boldsymbol{w}$  is null if $i$ equals $j$, and the model will predict whether some words need to be inserted in position $i$.}
\paragraph{TS with Hints.}
In the sub-task \emph{TS with hints}, the hints of translators are considered as some soft constraints for the model, and the model is expected to generate suggestions meeting these constraints. The format of the translator's hint is very flexible, which usually requires only a few types on the keyboard by the translator. For English and German, the hints can be the character sequence which includes the initials of words in the correct alternative. As for Chinese, the hints can be the character sequence which includes the initials of the phonetics of words in the correct alternative. In this setting, the model is optimized as:
\begin{equation}
    P(\boldsymbol{y}|\boldsymbol{x}, \boldsymbol{m}^{-\boldsymbol{w}}, \boldsymbol{h}, \theta)
\end{equation}
where $\boldsymbol{h}$ indicates the hints provided by translators.

\paragraph{Related tasks.}
Some similar techniques have been explored in CAT. \citet{green2014human} and \citet{knowles2016neural} study the task of so-called translation prediction, which provides predictions of the next word (or phrase) given a prefix. \citet{huang2015new} and \citet{santy2019inmt} further consider the hints of the translator in the task of translation prediction. Compared to TS, the most significant difference is the strict assumption of the translation context, i.e., the prefix context, which severely impedes the use of their methods under the scenarios of PE. Lexically constrained decoding which completes a translation based on some unordered words, relaxes the constraints provided by human translators from prefixes to general forms \cite{hokamp2017lexically,post2018fast,kajiwara2019negative,susanto2020lexically}. Although it does not need to re-train the model, its low efficiency makes it only applicable in scenarios where only a few constraints need to be applied. Recently, \citet{li2021gwlan} study the problem of auto-completion with different context types. However, they only focus on the word-level auto-completion, and their experiments are also conducted on the automatically constructed datasets.

\section{Data Description}
This section introduces the proposed dataset \emph{WeTS} used in the shred task, which is a golden corpus for four translation directions, including English-to-German, German-to-English, Chinese-to-English and English-to-Chinese.

\begin{table}[ht]
			\centering
			\resizebox{0.95 \columnwidth}{!}{
				\begin{tabular}{c|ccc}
					\toprule[2pt]
					 Translation Direction & Train & Valid & Test \\
					\midrule[1pt]
			        En$\Rightarrow$De & 14,957 & 1000 & 1000 \\
			        De$\Rightarrow$En & 11,777& 1000 & 1000 \\
			        Zh$\Rightarrow$En & 21,213& 1000 & 1000  \\
			        En$\Rightarrow$Zh & 15,769& 1000 & 1000  \\
					\bottomrule[2pt]
				\end{tabular}}
					\caption{\label{tab:statics of WeTs} The sizes for cases in train/valid/test sets. ``En$\Rightarrow$De” refers to the direction of English-to-German, and ``En$\Rightarrow$Zh” refers to English-to-Chinese.}
\end{table}
\begin{figure*}[t]
    \centering
    \includegraphics[scale=0.85]{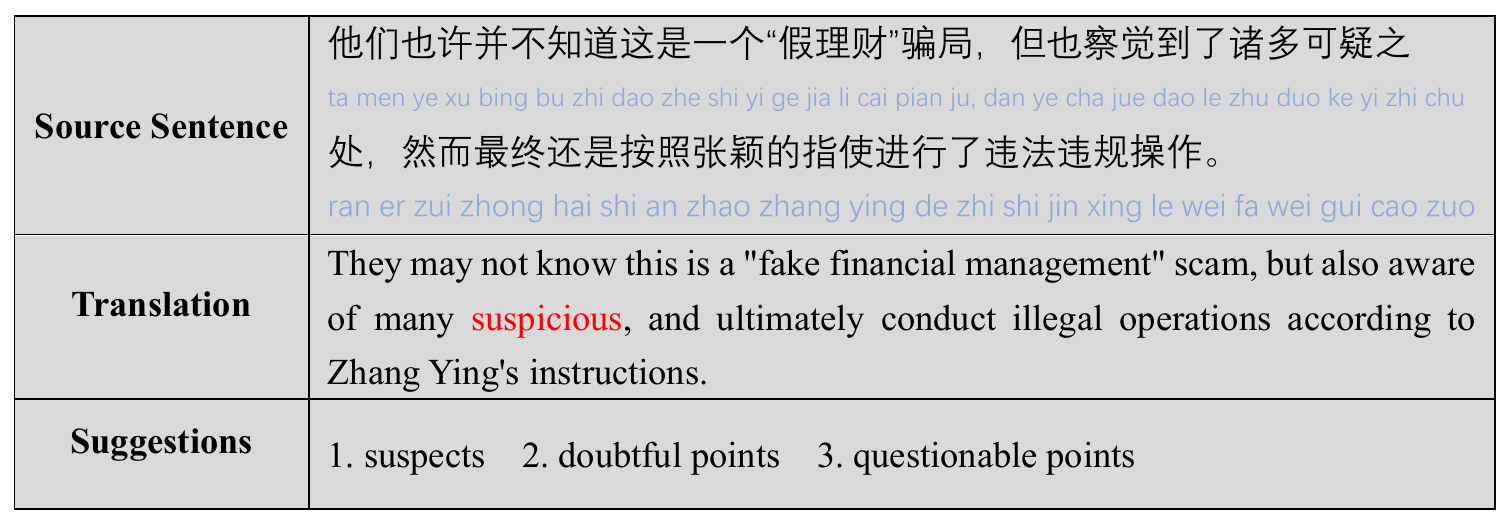}
    \caption {One training example in \emph{WeTS}. For the incorrect word "suspicious" (in red color), there are three correct suggestions. For readability, we also provide the Chinese pinyin format for the Chinese sentence (in blue color). }
    \label{fig:example}
\end{figure*}

\begin{figure*}
    \centering
    \includegraphics[scale=0.45]{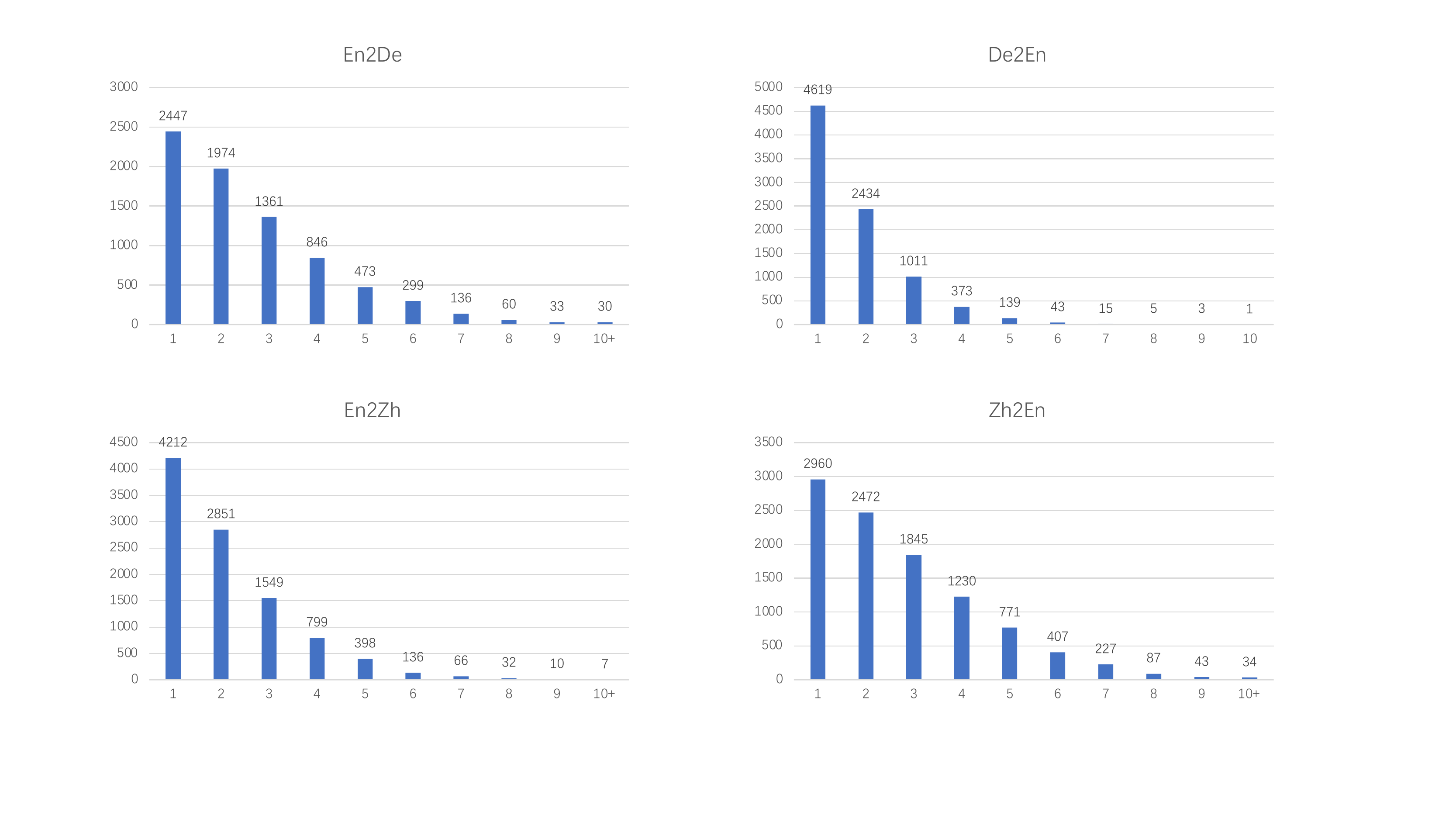}
    \caption {The number of incorrect span in each annotated example.}
    \label{fig:n_span}
\end{figure*}

\begin{figure*}
    \centering
    \includegraphics[scale=0.45]{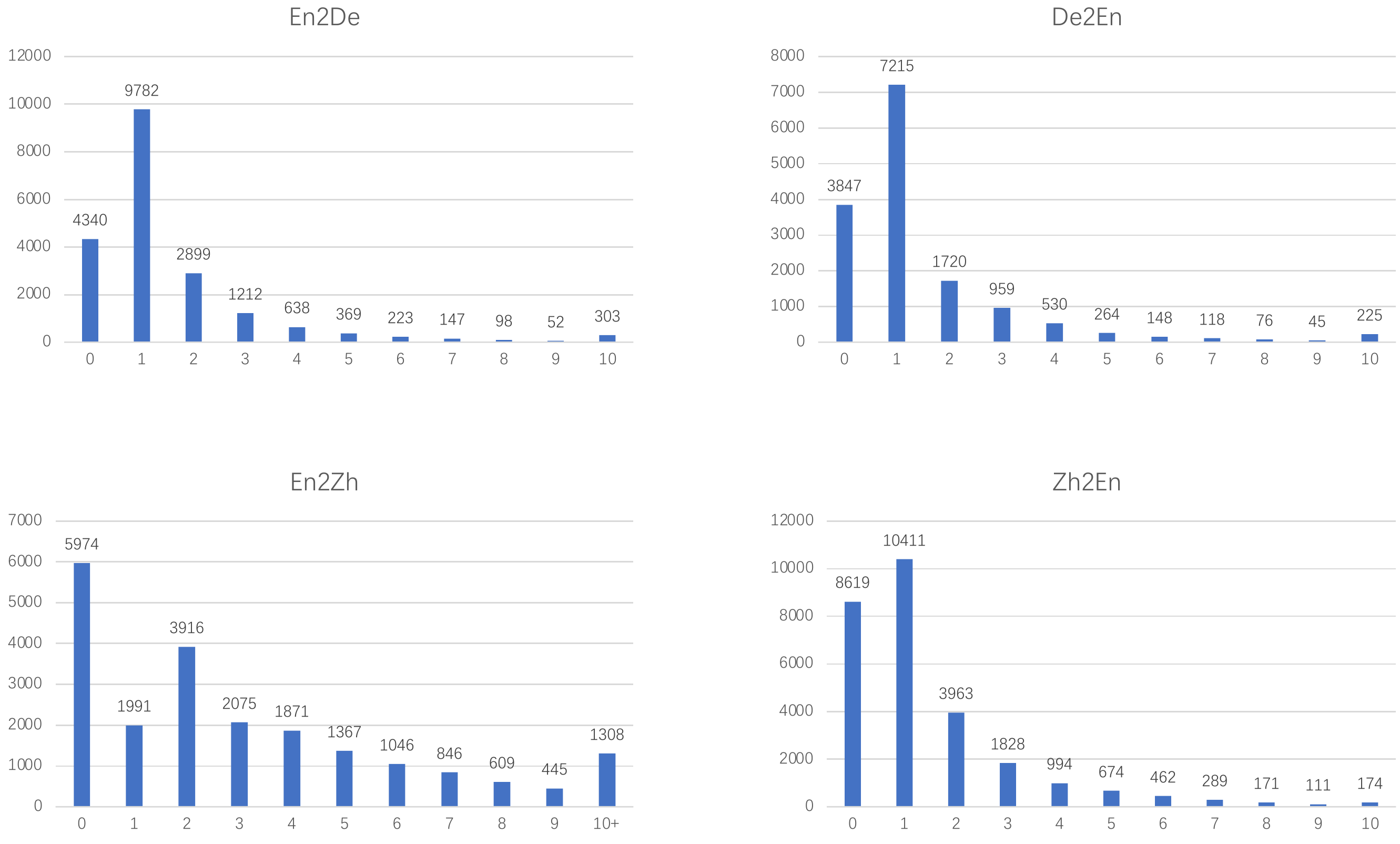}
    \caption {The length of the incorrect span.}
    \label{fig:l_span}
\end{figure*}

\subsection{Annotation Guidelines}
It is non-trivial for annotators to locate the incorrect word spans in the MT sentence. The main difficulty is that, the concept of ``translation error" is ambiguous and each translator has his own understanding about translation errors. To easier the annotation workload and reduce the possibility of making errors, we group the translation errors on which we aim to focus into three macro categories:
%all annotators are asked to work in accordance with the following annotation guidelines:
\begin{itemize}
    \item Under-translation or over-translation: While the problem of under-translation or over-translation has been alleviated with the popularity of Transformer, it is still one of the main mistakes in NMT and seriously destroys the readability of the translation.
    \item Semantic errors:  For the semantic error, we mean that some source words are incorrectly translated according to the semantic context, such as the incorrect translations for entities, proper nouns, and ambiguous words. Another branch of semantic mistake is that the source words or phrases are only translated superficially and the semantics behind are not translated well.
    \item Grammatical or syntactic errors: Such errors usually appear in translations of long sentences, including the improper use of tenses, passive voice, syntactic structures, etc.
\end{itemize}
 Another key rule for translators is that annotating the incorrect span as local as possible, as generating correct alternatives for long sequences is much harder than that of shorter sequences. 
 
\subsection{Data Construction}
As the starting point, we collect the monolingual corpora for English and German from the raw Wikipedia dumps, and extract Chinese monolingual corpus from various online news publications. We first clean the monolingual corpora with a language detector to remove sentences belonging to other languages.\footnote{\url{https://github.com/Mimino666/langdetect}} For all monolingual corpora, we remove sentences that are shorter than 20 words or longer than 80 words. In addition, sentences which exist in the available parallel corpora are also removed. Then, we get the translations by feeding the cleaned monolingual corpus into the corresponding fully-trained NMT model.  The NMT models for English-German language pairs are trained on the parallel corpus of WMT14 English-German. For Chinese-English directions, the NMT models are trained with the combination between the WMT19 English-Chinese\footnote{\url{https://www.statmt.org/wmt19/translation-task.html}} and the same amount of in-house corpus. \footnote{We have released the models and inference scripts utilized here to make our results easy reproduced.}
%\footnote{To make our work more transparent to the followers, we will release the models we utilized here even it is not necessary for the followers.} 

Finally, the translators are required to mark the incorrect word spans in the translation sentence and 
provide at least one alternative for each incorrect span, by using the annotation guidelines. The team is composed by eight annotators with high expertise in translation and each example has been assigned to three experts. There are two phases of agreement computations. In the first phase,
an annotation is considered in agreement among the experts if and only if they capture the same incorrect word spans. If one annotation passes the first agreement computation, it will be assigned to other three experts in charge of selecting the right alternatives from the previous annotation. In the second phase of agreement computation, an annotation is considered in agreement among the experts if and only if they select the same right alternatives. With the two-phase agreement checking, we ensure the high quality of the annotated examples. For the annotated examples with multiple incorrect word spans, we can extract multiple examples which have the same source and translation sentences, but different incorrect word span and the corresponding suggestions. Finally the extracted examples are randomly shuffled and then split into the training, validation and test sets.\footnote{To keep the fairness of \emph{WeTS}, we ensure the examples among the training, validation and test sets have different source and translation sentences.} One training example for the translation direction of Chinese-to-English is presented in Figure \ref{fig:example} and the sizes for the train/valid/test sets in \emph{WeTS} are collected in Table \ref{tab:statics of WeTs}. 

\begin{figure*}
    \centering
    \includegraphics[scale=0.45]{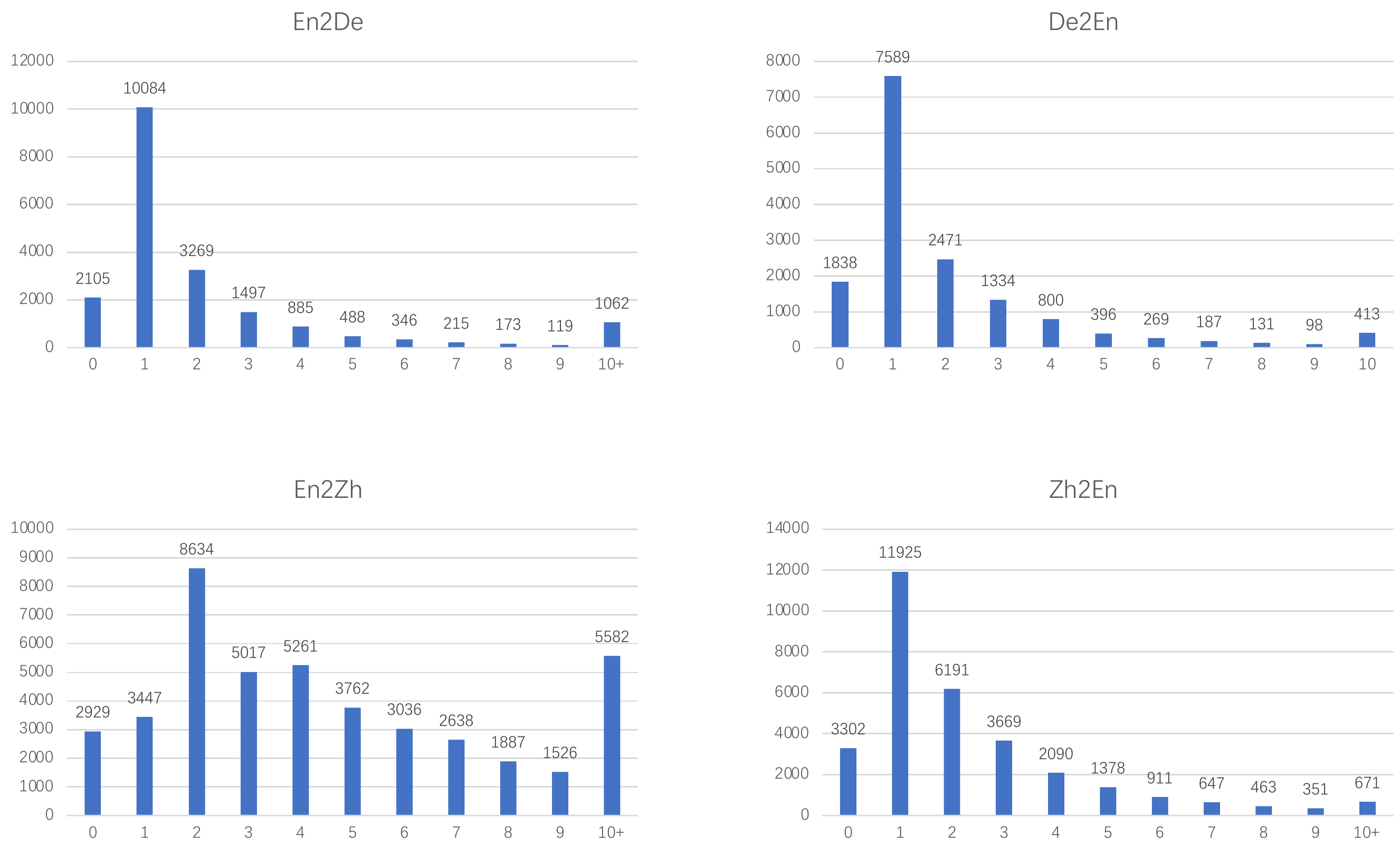}
    \caption {The length of the suggestion.}
    \label{fig:l_sugg}
\end{figure*}

\subsection{Detailed Statistics}
\paragraph{The number of the incorrect span}
Each annotated example may contain multiple incorrect spans, we show the number of the incorrect span in each annotated example as Figure \ref{fig:n_span}. We can see that most examples have only a few incorrect spans, and there are more than 70 percent examples containing less than 3 incorrect spans for each translation direction.  

\paragraph{The length of the incorrect span}
Figure \ref{fig:l_span} represents the length distribution of the incorrect spans. We can find that most of the incorrect spans contain less than 3 words or Chinese characters. This is mainly because of our key rule for annotating the incorrect span as local as possible. Additionally, for all of the four translation directions, the number of the incorrect spans with length 0 ranks top-2 among all the length buckets. This shows that under-translation is still a frequent error of the existing NMT models. 

\paragraph{The length of the suggestions}
Figure \ref{fig:l_sugg} shows the length distribution of the suggestions. We can see that in English-to-German, German-to-English and Chinese-to-English, most of the suggestions contain only one word. For English-to-Chinese, most suggestions contain two Chinese characters. Additionally, we can also find that there are quite a few of suggestions with length zero in each translation direction. This shows that over-translation is a non-negligible problem for the existing NMT models.

\section{Participants}
Five participants submitted their systems to the sub-task one of TS shared task. And two participants submitted their systems to the second sub-task. In sub-task one, 92 runs were submitted in total (each team is only allowed to submit less than 15 runs). Table \ref{tab:teams} summarizes the participants and their affiliations.
\begin{table}[ht]
			\centering
			\resizebox{0.95 \columnwidth}{!}{
				\begin{tabular}{cc}
					\toprule[2pt]
					 Team & Institution \\
					\midrule[1pt]
			        mind-ts & Soochow University and Alibaba \\
			        suda-hlt & Soochow University \\
			        Avocados & Beijing Jiaotong University  \\
			        IOL Research & Transn IOL Technology CO., Ltd.  \\
			        Slack & Zhejiang University \\
					\bottomrule[2pt]
				\end{tabular}}
					\caption{\label{tab:teams} The participating teams and their affiliations.}
\end{table}

\subsection{Systems}
Here we briefly describe each participant's systems as described by the authors and refer the reader to the participant's submission for further details. Since some participants did not submit their papers, we only describe the systems in the submitted papers.
\subsubsection{Baseline}
We take the naive Transformer-base \cite{vaswani2017attention} as the baseline and directly apply the implementation of the open-source toolkit, fiarseq.\footnote{\url{https://github.com/pytorch/fairseq}} We construct the synthetic corpus based on the WMT parallel corpus, and we refer the readers for details about constructing the synthetic corpus in the paper \cite{yang2021wets}. For training, we apply the two-state training pipeline, where we pre-train the model on the synthetic corpus in the first stage, and then fine-tune the model on the golden corpus in the second stage.

\subsubsection{IOL Research}
The team of IOL Research participates the two sub-tasks and focuses on the En-Zh and Zh-En translation directions. They use the $\Delta$LM as their backbone model.
$\Delta$LM is a pre-trained multilingual encoder-decoder model, which outperforms various strong baselines on both natural language generation and translation tasks \cite{ma2021deltalm}. Its encoder and decoder are initialized with the pre-trained multilingual encoder InfoXLM \cite{chi2020infoxlm}. Their model has 360M parameters, 12-6 encoder-decoder layers, 768 hidden size, 12 attention heads and 3072 FFN dimension. For the training data, they construct the synthetic data with two different methods according to its constructing complexity. During training, they use the two-stage fine-tuning, where they apply the synthetic data to fine-tune the original $\Delta$LM in the first stage and then fine-tune the result of the first stage with the golden corpus. In their experiments, they find that the accuracy indicator of TS can be helpful for efficient PE in practice. Overall, they achieved the best scores on 3 tracks and comparable result on another track.
\subsubsection{Avocados}
The team of Avocados tries different model structures, such as Transformer-base \cite{vaswani2017attention}, Transformer-big \cite{vaswani2017attention}, SA-Transformer \cite{yang2021wets} and DynamicConv \cite{wu2019pay}. They test different ensemble approaches for better performance. 
For more details, we refer the readers to their paper \cite{zhang2022improved}. Their main efforts are paid on building the synthetic corpus. They apply three different ways to construct the synthetic corpus. Firstly, they randomly sample a sub-segment in each target sentence of the golden parallel data, mask the sampled sub-segment to simulate an incorrect span, and use the sub-segment as an alternative suggestion. Secondly, the same strategy as above is used for pseudo-parallel data with the target side substituted by machine translation results. Finally, they use a quality estimation model to estimate the translation quality of words in translation output sentence and select the span with low confidence for masking. Then, an alignment tool to find the sub-segment corresponding to the span in the reference sentence and use it as the alternative suggestion for the span. To bridge the domain difference between the large-scale synthetic data and human-annotated golden corpus, they apply the pre-trained BERT to filter data similar to the golden corpus as in-domain data, which are used as pre-training for the next phase after pre-training model with a large-scale synthetic corpus. Overall, they rank second and third on the English-German and English-Chinese bidirectional tasks respectively.
\subsubsection{mind-ts}
The team of mind-ts participate in the English-German and English-Chinese translation directions in the sub-task one, and their submissions are ranked first in three of four language directions. For English-German, they initialize the weights with NMT models released by teh winner of WMT19 \cite{ng2019facebook}. For English-Chinese, the one-to-many and many-to-one mBART50 models are used \cite{tang2020multilingual}. Their main contribution is to construct the synthetic corpus with word alignment. They use the well-trained alignment models between source and target languages to filter out high-quality augment data. Specifically, they first use the Fast Align toolkit to extract the token alignments. Then, they remove tokens that appear in both MT and reference to get the trimmed result. They trim these common tokens because they want the model to focus more on the incorrect span and its alternative. Additionally, they use the dual conditional cross-entropy model to calculate the quality score of the pair between the source and masked translation sentences. If the cross-entropy quality score meets the threshold, they treat the masked translation and the alignment segments as the good examples for TS. Similarly, they also use the two-phase pre-training pipeline to 
get the final models.
\subsection{Submission Summary}
The submissions for this year's TS shared task cover different approaches from the pre-trained LMs and the encoder-decode NMT models. From the submissions, we find that the pre-trained models are very useful for the final performance. Additionally, almost all of the submissions have tried different approaches for constructing the synthetic corpus. As the amount of the golden corpus is limited, it is very important to find efficient ways to construct the synthetic corpus. The main problem for constructing synthetic corpus is how to make the synthetic corpus similar to the golden corpus in domain or other aspects. Finally, how to efficiently apply the synthetic corpus also needs much more efforts to investigate. All submissions adopt the two-stage training pipeline to train the models. 

\subsection{Evaluation Results}
We report the BLEU scores of the submissions. The BLEU is calculated automatically with the sacrebleu toolkit. For each run, the participating team need to submit their top-1 suggestions for each sentence in the test set. Each participating team can submit at most 15 times for each track. We only report the best score for each team. Table \ref{tab:en-zh-results} and \ref{tab:en-de-results} report the results on English-Chinese and English-German respectively in the sub-task one. Table \ref{tab:en-zh-results-2} report the results on English-Chinese in the sub-task two.

\begin{table}[ht]
			\centering
			\resizebox{0.80 \columnwidth}{!}{
				\begin{tabular}{ccc}
					\toprule[2pt]
					 Team & En-Zh & Zh-En \\
					\midrule[1pt]
					Baseline & 31.02 & 25.84\\
					\hline \hline
			        mind-ts & 33.92(2) & 30.07 (1) \\
			        Avocados & 33.33 (3) & 28.56 (3) \\
			        IOL Research & 39.71 (1) & 28.42 (4) \\
					\bottomrule[2pt]
				\end{tabular}}
					\caption{\label{tab:en-zh-results} Evaluation results on the language pair for English-Chinese in the sub-task one. The number in bracket is the ranked position.}
\end{table}

\begin{table}[ht]
			\centering
			\resizebox{0.80 \columnwidth}{!}{
				\begin{tabular}{ccc}
					\toprule[2pt]
					 Team & En-De & De-En \\
					\midrule[1pt] 
					Baseline & 35.07 & 37.61\\
					 \hline \hline
			        mind-ts & 42.91(1) & 47.04 (1) \\
			        Avocados & 42.61 (2) & 36.30 (2) \\
					\bottomrule[2pt]
				\end{tabular}}
					\caption{\label{tab:en-de-results} Evaluation results on the language pair for English-German in the sub-task one. The number in bracket is the ranked position.}
\end{table}

\begin{table}[ht]
			\centering
			\resizebox{0.80 \columnwidth}{!}{
				\begin{tabular}{ccc}
					\toprule[2pt]
					 Team & En-Zh & Zh-En \\
					\midrule[1pt]
					Baseline & 41.83 & 35.02\\
					\hline \hline
			        IOL Research & 48.60 (1) & 39.95 (1) \\
					\bottomrule[2pt]
				\end{tabular}}
					\caption{\label{tab:en-zh-results-2} Evaluation results on the language pair for English-Chinese in the sub-task two. The number in bracket is the ranked position.}
\end{table}

\section{Discussion and Analysis}
Comparing the results of the BLEU scores of all submissions with our baseline systems, there is a significant gap between the submitted and baseline systems.
This shows that there is a large space for us to try different techniques to improve the performance of TS. By comparing the results of different submitted systems, we find that different pre-training models have a large difference on the final performance. This is a similar trend with other NLP tasks. Therefore, we believe that this is an interesting and promising direction for us to pay much more efforts.

All submitted systems have investigated different approaches for constructing the synthetic corpus and almost all of them have achieved much improvements with the synthetic corpus. The noise in the synthetic corpus is a major problem which negatively affects the final performance. Therefore, how to filter or decrease the noise is an open question. The team of mind-ts applies the pre-trained LM to filter the synthetic corpus and obtain better performance on 3 out of 4 tracks based on the high-quality synthetic corpus. We can investigate more effective approaches to detect and filter the noise in the synthetic corpus.

However, another interesting direction which are not investigated by the submissions is modeling the interaction between the source and translation sentences efficiently. Compared to MT, the main difference for TS is that the input for TS is dual-source, namely the source and translation sentence. We believe that efficiently modeling the interaction between the source and translation sentences can improve the final performance.

\section{Conclusion}
We present the results of first edition of the Translation Suggestion shared task. For the goal of this task, we create and release the first golden benchmark dataset, called \emph{WeTS}, which covers the language pairs for English-Chinese and English-German. We wish the released corpus can spur the researches in this area. This year we received 92 submissions from 5 participating teaming in the sub-task one and 6 submissions for the sub-task 2, most of them covering the two translation directions. Results of these submissions show that the pre-trained models and synthetic corpus are two important factors for the final performance.

\section*{Acknowledgements}
We would like to thank Yaou Li and Ning Zhang for their helps on building the official website of the shared task. The authors would also like to thank the anonymous reviewers of this paper, and the anonymous reviewers of the previous version for their valuable comments and suggestions to improve our work.

\bibliography{anthology,custom}

\begin{thebibliography}{27}
\expandafter\ifx\csname natexlab\endcsname\relax\def\natexlab#1{#1}\fi

\bibitem[{Alabau et~al.(2014)Alabau, Buck, Carl, Casacuberta,
  Garc{\'\i}a-Mart{\'\i}nez, Germann, Gonz{\'a}lez-Rubio, Hill, Koehn, Leiva
  et~al.}]{alabau2014casmacat}
Vicent Alabau, Christian Buck, Michael Carl, Francisco Casacuberta, Mercedes
  Garc{\'\i}a-Mart{\'\i}nez, Ulrich Germann, Jes{\'u}s Gonz{\'a}lez-Rubio,
  Robin Hill, Philipp Koehn, Luis~A Leiva, et~al. 2014.
\newblock Casmacat: A computer-assisted translation workbench.
\newblock In \emph{Proceedings of the Demonstrations at the 14th Conference of
  the European Chapter of the Association for Computational Linguistics}, pages
  25--28.

\bibitem[{Bahdanau et~al.(2015)Bahdanau, Cho, and Bengio}]{bahdanau:14}
Dzmitry Bahdanau, Kyunghyun Cho, and Yoshua Bengio. 2015.
\newblock Neural machine translation by jointly learning to align and
  translate.
\newblock In \emph{3rd International Conference on Learning Representations,
  {ICLR} 2015, San Diego, CA, USA, May 7-9, 2015, Conference Track
  Proceedings}.

\bibitem[{Barrachina et~al.(2009)Barrachina, Bender, Casacuberta, Civera,
  Cubel, Khadivi, Lagarda, Ney, Tom{\'a}s, Vidal
  et~al.}]{barrachina2009statistical}
Sergio Barrachina, Oliver Bender, Francisco Casacuberta, Jorge Civera, Elsa
  Cubel, Shahram Khadivi, Antonio Lagarda, Hermann Ney, Jes{\'u}s Tom{\'a}s,
  Enrique Vidal, et~al. 2009.
\newblock Statistical approaches to computer-assisted translation.
\newblock \emph{Computational Linguistics}, 35(1):3--28.

\bibitem[{Chi et~al.(2020)Chi, Dong, Wei, Yang, Singhal, Wang, Song, Mao,
  Huang, and Zhou}]{chi2020infoxlm}
Zewen Chi, Li~Dong, Furu Wei, Nan Yang, Saksham Singhal, Wenhui Wang, Xia Song,
  Xian-Ling Mao, Heyan Huang, and Ming Zhou. 2020.
\newblock Infoxlm: An information-theoretic framework for cross-lingual
  language model pre-training.
\newblock \emph{arXiv preprint arXiv:2007.07834}.

\bibitem[{Cho et~al.(2014)Cho, van Merrienboer, G{\"u}l{\c{c}}ehre, Bahdanau,
  Bougares, Schwenk, and Bengio}]{cho2014learning}
Kyunghyun Cho, Bart van Merrienboer, {\c{C}}aglar G{\"u}l{\c{c}}ehre, Dzmitry
  Bahdanau, Fethi Bougares, Holger Schwenk, and Yoshua Bengio. 2014.
\newblock Learning phrase representations using rnn encoder-decoder for
  statistical machine translation.
\newblock In \emph{EMNLP}.

\bibitem[{Green et~al.(2013)Green, Heer, and Manning}]{green2013efficacy}
Spence Green, Jeffrey Heer, and Christopher~D Manning. 2013.
\newblock The efficacy of human post-editing for language translation.
\newblock In \emph{Proceedings of the SIGCHI conference on human factors in
  computing systems}, pages 439--448.

\bibitem[{Green et~al.(2014)Green, Wang, Chuang, Heer, Schuster, and
  Manning}]{green2014human}
Spence Green, Sida~I Wang, Jason Chuang, Jeffrey Heer, Sebastian Schuster, and
  Christopher~D Manning. 2014.
\newblock Human effort and machine learnability in computer aided translation.
\newblock In \emph{Proceedings of the 2014 Conference on Empirical Methods in
  Natural Language Processing (EMNLP)}, pages 1225--1236.

\bibitem[{Hokamp and Liu(2017)}]{hokamp2017lexically}
Chris Hokamp and Qun Liu. 2017.
\newblock \href {https://doi.org/10.18653/v1/P17-1141} {Lexically constrained
  decoding for sequence generation using grid beam search}.
\newblock In \emph{Proceedings of the 55th Annual Meeting of the Association
  for Computational Linguistics (Volume 1: Long Papers)}, pages 1535--1546,
  Vancouver, Canada. Association for Computational Linguistics.

\bibitem[{Huang et~al.(2015)Huang, Zhang, Zhou, and Zong}]{huang2015new}
Guoping Huang, Jiajun Zhang, Yu~Zhou, and Chengqing Zong. 2015.
\newblock A new input method for human translators: integrating machine
  translation effectively and imperceptibly.
\newblock In \emph{Twenty-Fourth International Joint Conference on Artificial
  Intelligence}.

\bibitem[{Kajiwara(2019)}]{kajiwara2019negative}
Tomoyuki Kajiwara. 2019.
\newblock Negative lexically constrained decoding for paraphrase generation.
\newblock In \emph{Proceedings of the 57th Annual Meeting of the Association
  for Computational Linguistics}, pages 6047--6052.

\bibitem[{Knowles and Koehn(2016)}]{knowles2016neural}
Rebecca Knowles and Philipp Koehn. 2016.
\newblock Neural interactive translation prediction.
\newblock In \emph{Proceedings of the Association for Machine Translation in
  the Americas}, pages 107--120.

\bibitem[{Lee et~al.(2021)Lee, Ahn, Park, and Jo}]{lee2021intellicat}
Dongjun Lee, Junhyeong Ahn, Heesoo Park, and Jaemin Jo. 2021.
\newblock \href {https://doi.org/10.18653/v1/2021.acl-demo.2} {{I}ntelli{CAT}:
  Intelligent machine translation post-editing with quality estimation and
  translation suggestion}.
\newblock In \emph{Proceedings of the 59th Annual Meeting of the Association
  for Computational Linguistics and the 11th International Joint Conference on
  Natural Language Processing: System Demonstrations}, pages 11--19, Online.
  Association for Computational Linguistics.

\bibitem[{Li et~al.(2021)Li, Liu, Huang, and Shi}]{li2021gwlan}
Huayang Li, Lemao Liu, Guoping Huang, and Shuming Shi. 2021.
\newblock \href {https://doi.org/10.18653/v1/2021.acl-long.370} {{GWLAN}:
  General word-level {A}utocompletio{N} for computer-aided translation}.
\newblock In \emph{Proceedings of the 59th Annual Meeting of the Association
  for Computational Linguistics and the 11th International Joint Conference on
  Natural Language Processing (Volume 1: Long Papers)}, pages 4792--4802,
  Online. Association for Computational Linguistics.

\bibitem[{Ma et~al.(2021)Ma, Dong, Huang, Zhang, Muzio, Singhal, Awadalla,
  Song, and Wei}]{ma2021deltalm}
Shuming Ma, Li~Dong, Shaohan Huang, Dongdong Zhang, Alexandre Muzio, Saksham
  Singhal, Hany~Hassan Awadalla, Xia Song, and Furu Wei. 2021.
\newblock Deltalm: Encoder-decoder pre-training for language generation and
  translation by augmenting pretrained multilingual encoders.
\newblock \emph{arXiv preprint arXiv:2106.13736}.

\bibitem[{Ng et~al.(2019)Ng, Yee, Baevski, Ott, Auli, and
  Edunov}]{ng2019facebook}
Nathan Ng, Kyra Yee, Alexei Baevski, Myle Ott, Michael Auli, and Sergey Edunov.
  2019.
\newblock Facebook fair's wmt19 news translation task submission.
\newblock \emph{arXiv preprint arXiv:1907.06616}.

\bibitem[{Papineni et~al.(2002)Papineni, Roukos, Ward, and
  Zhu}]{papineni2002bleu}
Kishore Papineni, Salim Roukos, Todd Ward, and Wei-Jing Zhu. 2002.
\newblock Bleu: a method for automatic evaluation of machine translation.
\newblock In \emph{Proceedings of the 40th annual meeting of the Association
  for Computational Linguistics}, pages 311--318.

\bibitem[{Post(2018)}]{post-2018-call}
Matt Post. 2018.
\newblock \href {https://www.aclweb.org/anthology/W18-6319} {A call for clarity
  in reporting {BLEU} scores}.
\newblock In \emph{Proceedings of the Third Conference on Machine Translation:
  Research Papers}, pages 186--191, Belgium, Brussels. Association for
  Computational Linguistics.

\bibitem[{Post and Vilar(2018)}]{post2018fast}
Matt Post and David Vilar. 2018.
\newblock \href {https://doi.org/10.18653/v1/N18-1119} {Fast lexically
  constrained decoding with dynamic beam allocation for neural machine
  translation}.
\newblock In \emph{Proceedings of the 2018 Conference of the North {A}merican
  Chapter of the Association for Computational Linguistics: Human Language
  Technologies, Volume 1 (Long Papers)}, pages 1314--1324, New Orleans,
  Louisiana. Association for Computational Linguistics.

\bibitem[{Santy et~al.(2019)Santy, Dandapat, Choudhury, and
  Bali}]{santy2019inmt}
Sebastin Santy, Sandipan Dandapat, Monojit Choudhury, and Kalika Bali. 2019.
\newblock Inmt: Interactive neural machine translation prediction.
\newblock In \emph{Proceedings of the 2019 Conference on Empirical Methods in
  Natural Language Processing and the 9th International Joint Conference on
  Natural Language Processing (EMNLP-IJCNLP): System Demonstrations}, pages
  103--108.

\bibitem[{Susanto et~al.(2020)Susanto, Chollampatt, and
  Tan}]{susanto2020lexically}
Raymond~Hendy Susanto, Shamil Chollampatt, and Liling Tan. 2020.
\newblock \href {https://doi.org/10.18653/v1/2020.acl-main.325} {Lexically
  constrained neural machine translation with {L}evenshtein transformer}.
\newblock In \emph{Proceedings of the 58th Annual Meeting of the Association
  for Computational Linguistics}, pages 3536--3543, Online. Association for
  Computational Linguistics.

\bibitem[{Tang et~al.(2020)Tang, Tran, Li, Chen, Goyal, Chaudhary, Gu, and
  Fan}]{tang2020multilingual}
Yuqing Tang, Chau Tran, Xian Li, Peng-Jen Chen, Naman Goyal, Vishrav Chaudhary,
  Jiatao Gu, and Angela Fan. 2020.
\newblock Multilingual translation with extensible multilingual pretraining and
  finetuning.
\newblock \emph{arXiv preprint arXiv:2008.00401}.

\bibitem[{Vaswani et~al.(2017)Vaswani, Shazeer, Parmar, Uszkoreit, Jones,
  Gomez, Kaiser, and Polosukhin}]{vaswani2017attention}
Ashish Vaswani, Noam Shazeer, Niki Parmar, Jakob Uszkoreit, Llion Jones,
  Aidan~N Gomez, {\L}ukasz Kaiser, and Illia Polosukhin. 2017.
\newblock Attention is all you need.
\newblock In \emph{Advances in neural information processing systems}, pages
  5998--6008.

\bibitem[{Wang et~al.(2020)Wang, Zhang, Liu, Huang, and Zong}]{wang2020touch}
Qian Wang, Jiajun Zhang, Lemao Liu, Guoping Huang, and Chengqing Zong. 2020.
\newblock Touch editing: A flexible one-time interaction approach for
  translation.
\newblock In \emph{Proceedings of the 1st Conference of the Asia-Pacific
  Chapter of the Association for Computational Linguistics and the 10th
  International Joint Conference on Natural Language Processing}, pages 1--11.

\bibitem[{Wu et~al.(2019)Wu, Fan, Baevski, Dauphin, and Auli}]{wu2019pay}
Felix Wu, Angela Fan, Alexei Baevski, Yann~N Dauphin, and Michael Auli. 2019.
\newblock Pay less attention with lightweight and dynamic convolutions.
\newblock \emph{arXiv preprint arXiv:1901.10430}.

\bibitem[{Yang et~al.(2021)Yang, Zhang, Li, Meng, and Zhou}]{yang2021wets}
Zhen Yang, Yingxue Zhang, Ernan Li, Fandong Meng, and Jie Zhou. 2021.
\newblock Wets: A benchmark for translation suggestion.
\newblock \emph{arXiv preprint arXiv:2110.05151}.

\bibitem[{Zhang et~al.(2022)Zhang, Lai, Zhang, Huang, Chen, Xu, and
  Liu}]{zhang2022improved}
Hongxiao Zhang, Siyu Lai, Songming Zhang, Hui Huang, Yufeng Chen, Jinan Xu, and
  Jian Liu. 2022.
\newblock Improved data augmentation for translation suggestion.
\newblock \emph{arXiv preprint arXiv:2210.06138}.

\bibitem[{Zouhar et~al.(2021)Zouhar, Popel, Bojar, and
  Tamchyna}]{zouhar2021neural}
Vil{\'e}m Zouhar, Martin Popel, Ond{\v{r}}ej Bojar, and Ale{\v{s}} Tamchyna.
  2021.
\newblock \href {https://doi.org/10.18653/v1/2021.emnlp-main.801} {Neural
  machine translation quality and post-editing performance}.
\newblock In \emph{Proceedings of the 2021 Conference on Empirical Methods in
  Natural Language Processing}, pages 10204--10214, Online and Punta Cana,
  Dominican Republic. Association for Computational Linguistics.

\end{thebibliography}
\bibliographystyle{acl_natbib}

 \end{CJK} 
\end{document}